\documentclass[11pt,a4paper]{article}
\usepackage[hyperref]{acl2020}
\usepackage{times}
\usepackage{latexsym}

\usepackage{amsmath,amssymb,amsfonts}
\usepackage{algorithmic}
\usepackage{graphicx}
\usepackage{textcomp}
\usepackage{xcolor}
\usepackage{comment}

\usepackage{microtype}

\aclfinalcopy 

\title{The Inception Team at NSURL-2019 Task 8: Semantic Question Similarity in Arabic}

\author{Hana Al-Theiabat and
  Aisha Al-Sadi \thanks{* These authors contributed equally to the work}
    \\Jordan University of Science and Technology, Irbid, Jordan \\
  haaltheiabat13@cit.just.edu.jo,  asalsadi16@cit.just.edu.jo
  }

\date{}

\begin{document}
\maketitle
\begin{abstract}
This paper describes our method for the task of Semantic Question Similarity in Arabic in the workshop on NLP Solutions for Under Resourced Languages (NSURL). The aim is to build a model that is able to detect similar semantic questions in Arabic language for the provided dataset. Different methods of determining questions similarity are explored in this work. The proposed models achieved high F1-scores, which range from (88\% to 96\%). Our official best result is produced from the ensemble model of using pre-trained multilingual BERT model with different random seeds with 95.924\%  F1-Score, which ranks the first among nine participants teams.
\end{abstract}

\section{Introduction}

Semantic matching or semantic similarity is a significant part of natural language processing (NLP) field for its variety of tasks. It used to measure the similarity and the relationship between different textual elements, such as words, sentences, or documents. Semantic matching has been involved in many NLP applications; including question answering, where it is used to assess question answering and retrieval tasks by employing it to estimate the similarity of query answer among all candidate answers \cite{wang2016sentence}. In addition, it has played a significant role in top-k re-ranking in machine translation \cite{brown1993mathematics}, information extraction \cite{grishman1997information} and automatic text summarization \cite{ponzanelli2015summarizing}.

Natural language has complicated structures either from sequential or hierarchical perspectives, capturing the relationship between two questions is becoming a challenging task. For example, questions that have the same meaning while their words have a different order. An effective semantic matching algorithm, therefore, needs to consider an appropriate semantic representation to capture the similarity without being affected with words order. 

This paper focuses on detecting semantic question similarity, which is a common challenge in Question-and-answer (Q\&A) websites, such as Quora and Stack Overflow. This work targets Arabic questions dataset published by Mawdoo3 AI\footnote{ https://ai.mawdoo3.com/nsurl-2019-task8}. Most of these questions are related to information provided by Mawdoo3.com which is  the largest comprehensive Arabic content website.
For these websites, the benefit of detecting duplicate questions is to improve the efficiency of search engines by being aware of the different paraphrases of the same question.

The rest of paper is organized as follows. Section 2 presents related works. While Section 3 presents  some details about the dataset. Section 4, presents the proposed models for solving the semantic similarity in Arabic language task. Results for all proposed models and the final results are presented in Section 5. Finally, the paper conclusion is presented in Section 6.

\section{Related Work}
Semantic matching has been a long-established problem in NLP. Many approaches were proposed to solve this problem. The conventional approaches were mainly based on representing text as a vector of word features. The bag-of-words (BoW) method \cite{wu2008interpreting} employed the word occurrence and Term Frequency-Inverse Document Frequency (TF-IDF) \cite{paltoglou2010study}  as the word feature. However, these types of models disregard word meaning, orders, and even grammar. In contrast, word embedding models such as word2vec \cite{mikolov2013efficient} and Glove \cite{pennington2014glove} have been widely used instead of BoW as they can learn distributional semantic representation for words. So based on word embeddings, the Word Mover’s Distance (WMD) \cite{kusner2015word} was proposed to measure the dissimilarity between two texts assuming that similar words should have similar vectors. Although WMD can estimate semantic similarity between texts, the order, and interactions between words are excluded. 

Recently many deep learning models have been proposed for text matching. A common framework has been adopted is the Siamese architecture \cite{mueller2016siamese,pang2016text, severyn2015learning,wang2017bilateral} where the encoder, which can be either Convolutional Neural Network (CNN) or Recurrent Neural Network (RNN), is applied individually on the two input texts, so both texts are encoded into intermediate contextual representations. Then, the matching result is generated by performing a scoring mechanism over contextual representations. Although this framework supports parameter sharing in its network, it purely learns complicated relationships among texts.

Another framework is based on matching aggregation  \cite{wang2016compare} which first matches the small units (such as words) of two texts to produce comparison vectors, then these vectors are aggregated and fed into a CNN or RNN for the final classification. This framework improves capturing the interactive features between two texts, but still it limits exploring the matching in only word-word manner.

As the main focus of this paper is to detect semantically equivalent questions, the following is the review of related approaches that were adopted to detect duplicate questions on Quora dataset. As Quora recently published a dataset of 400K labeled questions, massive researches have been proposed on this dataset for question paraphrase identification challenge \cite{qourawebsite}.  One Relevant approach that was proposed for this challenge is the Bilateral Multi-Perspective Matching model (BIMPM) model  \cite{wang2017bilateral} which encodes two questions with a Bidirectional Long Short-Term Memory Network (BiLSTM). Then, a multi-perspective matching in the two directions is applied to both questions, and for each time step, questions are matched using different types of extensive matching. On Quora dataset, the result of this model reached  88.17\%.  In \cite{mirakyan2018natural}, a novel architecture can obtain a high-level understanding of the question pairs through extracting the semantic features using dense interaction tensors (attention) network which called Densely Interactive Inference Network (DIIN). DIIN outperforms BiLSTM on Quora to achieve accuracy of 89.06\%.  
Moreover, Multi-Task Deep Neural Network (MT-DNN) \cite{liu2019multi} achieved competitive performance on several tasks including question paraphrase on Quora with an accuracy of 89.6\%.  Specifically, MT-DNN Combined multi-task learning and pre-trained bidirectional transformer model for language representation learning. 

\section {Dataset Description}
The dataset used in this task is provided by Mawdoo3 ~\cite{Task8_SemQSimArabic}. It is a dataset for questions in Arabic language, it consists of 11,997 labeled question pairs as training data, and 3,715 question pairs as testing data. Label `1' means the question pairs are similar in semantic where label `0' means the opposite. 55\% of the training question pairs are with label `0', and 45\%  are with label `1'. The max length of question 1 is 14 words with an average of 5.7 words per question, while the max length of question 2 is 28 words with an average of 5.3 words per question. Table~\ref{datasetimg} shows samples from the training dataset.

\begin{table}[htb]

  \includegraphics[width= 0.47\textwidth]
{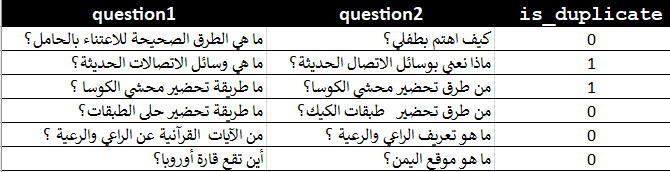}
  \caption{Question samples from Mawdoo3 dataset}
  \label{datasetimg}
  \end{table}

The only processing step that was applied to the dataset is to unify countries names, some examples are shown in Table~\ref{countrynames}.

\begin{table}[htb]
\centering
 
\includegraphics[width= 0.35\textwidth]
{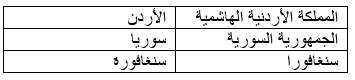}
 \caption{Unify countries names example}
  \label{countrynames}
  \end{table}

\section{Methodology}

In this work, four different deep learning approaches are presented to solve the semantic similarity task, which are RNN based model, CNN based model, multi-head attention based model, and finally BERT model. In this section, each model is discussed.

 \subsection{Convolutional Neural Network Model}
 In NLP field, CNN has shown the ability to extract most informative n-gram features from the input sequence, and then apply the activation on these features \cite{kim2014convolutional}. Although CNN is known for the applications in the image processing field, it is used here for text classification application.
 
\begin{figure}[htb]
\centering
\includegraphics[width= 0.48\textwidth]
{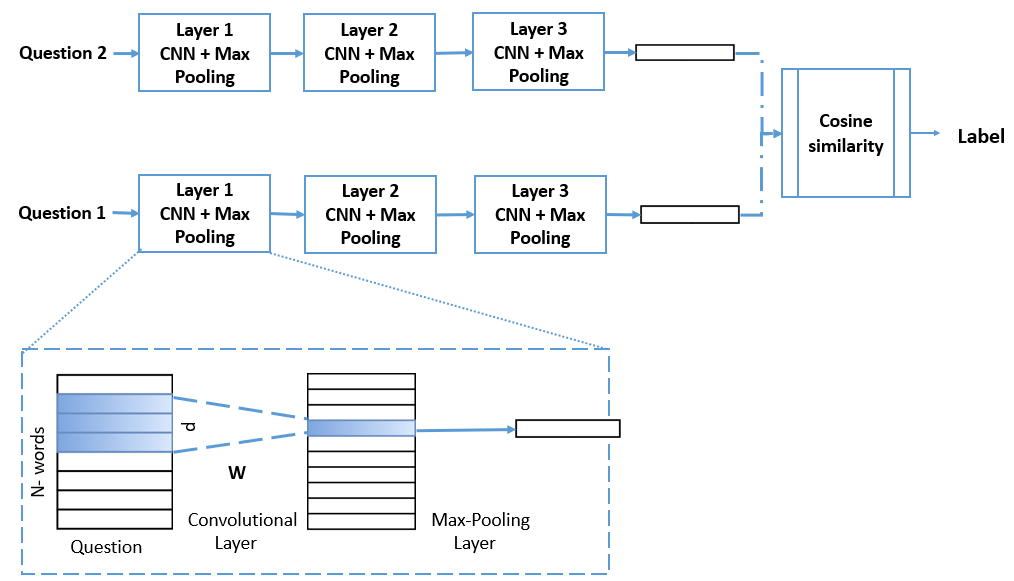}
\caption{CNN model architecture used for detecting semantic questions similarity}
\label{cnn_fig}
\end{figure}

The proposed model architecture is shown in Figure~\ref{cnn_fig}. Firstly, the words are mapped in the dictionary to get a representation for each word. Then each question is fed to three consecutive layers. In each layer, the convolutional layer is applied, followed by activation and then max pooling. Hence, each question's output is a feature representation which is used to get the similarity label by computing the cosine similarity between the two questions features.

 \subsection{Recurrent Neural Network Model}
 The significant advantage of RNNs is the computation of the same task over each element of the sequence, so the output for each block depends on the previous computations. Hence, RNN has been increasingly prevalent in NLP field specifically for RNN types that have a memory to remember the information through the sequence.

In this model, the input is a sequence of question-pairs that are concatenated to represent a single sequence. Then, the sequence is encoded by the dictionary to be fed into a bi-directional Gated Recurrent Units (GRUs) network with 128 hidden units to generate the similarity label as output. 

\vspace{0.4pt}
 \subsection{Multi-head Attention Network Model}
Multi-head attention model \cite{vaswani2017attention} allows to learn on various locations of the encoded words. Our network consists of a stacked encoder-decoder structure with eight heads. 

For each question-pairs of sequence length n, at each layer $l$, the encoder maps a sequence of words $ Q_l={w_1^l,..,w_n^l} $ into hidden representation $h^l={h_1^l,..,h_n^l}$. After computing the  attention on all positions jointly, the transformer stacks all hidden representation $h^l$ at the current layer $l$ together into matrix $H^l$.  Given $h$, the decoder then generates output sequence $ y^l={y_1^l,..,y_n^l} $, and after that apply softmax to estimate the output label. 
The transformer also contains two sub-layers, a multi-head attention layer, and a position-encoding layer.

The position-encoding layer benefits the network to keep track of relative positions for each word in the sequence since the context and the meaning of a sequence depend on the order of its words. 

In the multi-head attention layer, instead of computing single attention on the overall sequence, it jointly gets attention from different representations at different positions. As a result, each head looks differently on encoder output, and the decoder easily learns to retrieve valuable information from the encoder.


\vspace{0.4pt}
\subsection{BERT Model}

Recently, pre-training language models have shown a significant role to improve many NLP tasks including question-pairs paraphrasing \cite{dolan2005automatically}. There are two approaches to apply these pre-trained language representations on NLP tasks; either feature-based or fine-tuning. For the feature-based approach \cite{peters2018deep}, researchers use the output of pre-trained model as additional features in their models, based on the task they target. On the other hand, the fine-tuning approach \cite{radford2018improving} permits the model to be trained on another task by learning task-specific parameters. The two strategies were mentioned previously have limitations to learning general language representations since they adopt the left-to-right unidirectional architectures.  On the other hand, Bidirectional Encoder Representations from Transformers (BERT) \cite{devlin2018bert}, has  strongly outperformed previous cutting-edge unidirectional models.





BERT model relies on the multi-head self-attention mechanism, which enables it to achieve the state-of-the-art accuracy on a wide range of tasks such as, natural language inference, question answering, and sentence classification. The architecture of BERT model is built upon the transformer layer, which is called the self-attention layer. 
For each layer, the representations of words are exchanged from previous layers regardless of their positions, in contrast to traditional unidirectional models. For each input word, the model learns bidirectional encoder representations by using the masked language model, which randomly masks some of the words from the input to predict the masked word contextually.

\begin{figure}[htb]
\centering
\includegraphics[width= 0.45\textwidth]
{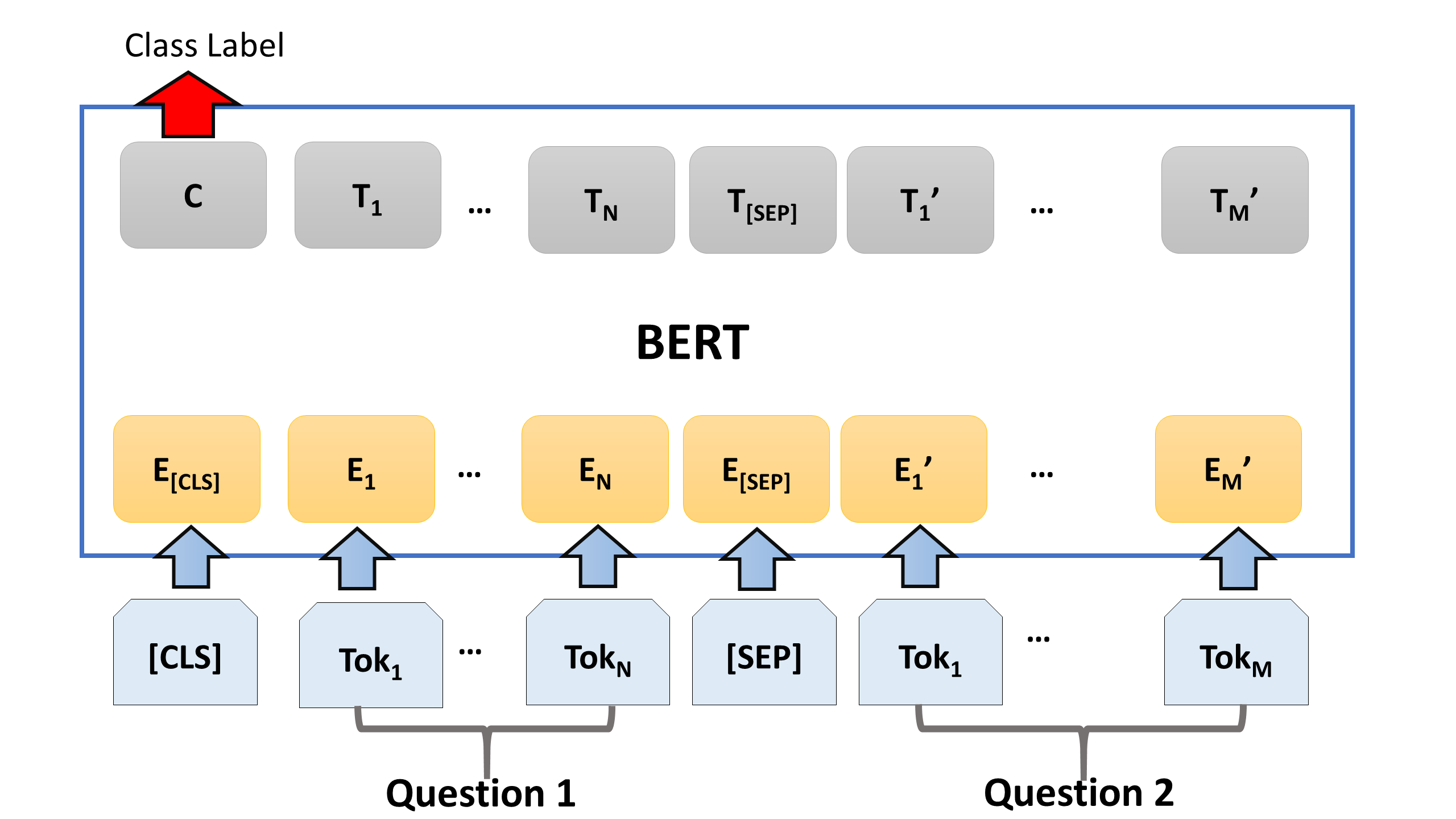}
\caption{BERT model used for question pair similarity classification task  
}
\label{bert}
\end{figure}

 As BERT offers pre-trained models for English language and multilingual model for 104 languages \cite{bertgithub} including the Arabic language, we applied the sentence pairs classification task on Arabic questions through fine-tuning the multilingual model as illustrated in Figure~\ref{bert}. 

\section{Experiments and Results}
For each of the four models explained in the methodology section, different hyper-parameters are used, such as learning rate, number of hidden nodes, and number of epochs. Table~\ref{tab:parameters} shows the main parameters values that give the best results for each model.


\begin{table}[htb]
\centering
\resizebox{\columnwidth}{!}{%
\begin{tabular}{|c|l|}
\hline
\multicolumn{1}{|c|}{Model}   & \multicolumn{1}{c|}{ Main parameters}     \\ \hline
 
              &  hidden size = 128   \\
             &  cell type = GRU \\ 
RNN Model      & bidirectional = true \\
              & number of train epochs = 10 \\
              & train batch size = 512 \\
              &learning rate = 0.001 \\ \hline

   &  number of filters = 50 , 50, 50  \\
              &   filter sizes = 2, 3, 4 \\
  CNN Model           & number of blocks = 2 \\ 
            & number of train epochs = 10 \\
              & train batch size = 512 \\
              &learning rate = 0.001 \\ \hline

                     & number of heads = 8 \\
                            & use residual = false \\
                            &  layer normalization = false \\ 
  Multi-Head Attention Model  & number of train epochs = 10 \\
              & train batch size = 512 \\
              &learning rate = 0.001 \\ \hline

             & max seq length = 50 \\
 BERT Model  & train batch size = 8 \\
             & learning rate = 2e-5 \\
             & number of train epochs=50 \\ \hline
   
\end{tabular}}
\caption{Main parameters for each proposed model}
\label{tab:parameters}
\end{table}

The evaluation metric that was used for this task is F1-Score that measures the precision $p$ and recall $r$ together as illustrated in the  equations [1]-[3]:

\begin{equation}
F1 = 2 p.r /(p  + r)
\end{equation}

\begin{equation}
p = tp / (tp+fp)
\end{equation}

\begin{equation}
r = tp /(tp + fn)
\end{equation}

where:
\begin{itemize}
    \item tp: true positive examples
    \item fp: false positive examples
    \item  fn: false negative examples
\end{itemize}

\vspace{0.2in}

Test data evaluation is automatically done online on Kaggle website by submitting the test predictions file. The evaluation system is as the following:
\begin{itemize}
    \item Public score: calculated with approximately  30\% of the data
    \item Private score: calculated with approximately 70\% of the data
\end{itemize}

During the competition, the public score for each submitted file was shown directly. Then after the competition ended, the submitted file with the highest public score was chosen to calculate its private score and compete other teams based on it.

Table~\ref{tab:test_results} shows the highest F1-Score for each of the four models for the public score and the private score of the test data. As illustrated, BERT model with pre-trained multilingual outperforms the remaining models with F1-score of 96.050\% on the public score, and 95.617\% on the private score.

\begin{table*}
\centering
\begin{tabular}{|l|c|c|}
\hline
\multicolumn{1}{|c|}{Model}   & \multicolumn{1}{c|}{ Public Score (\%)}  & \multicolumn{1}{c|}{ Private Score (\%)}   \\ \hline
 
 RNN Model    &    88.061      &  88.312  \\ \hline
 CNN Model    &    88.330     &  88.773  \\ \hline
 Multi-Head Attention Model    &  86.804 &  87.889 \\ \hline
 BERT Model  &   96.050 & 95.617  \\ \hline
   
\end{tabular}
\caption{Results of 30\% of the test data}
\label{tab:test_results}
\end{table*}

\begin{table*}
\centering
\begin{tabular}{|l|l|l|}
\hline
\multicolumn{1}{|c|}{Model}   & \multicolumn{1}{c|}{  Public Score  (\%)}  & \multicolumn{1}{c|}{ Private Score (\%)}   \\ \hline
 
  Ensemble of best 3 seeds  & 95.960  & 96.155  \\ \hline
  Ensemble of best 4 seeds  & 96.499  & 95.924  \\ \hline
  Ensemble of best 5 seeds  & 95.691  & 96.232 \\ \hline
  Ensemble of best 6 seeds  & 95.332  & 96.001  \\ \hline
 
\end{tabular}
\caption{BERT Ensemble Results}
\label{tab:bert_ensemble}
\end{table*}

Note that the previous results are based on the best public score for every single model of the four models. Since BERT model gives the best results, we conducted other experiments with different random seeds in order to ensemble BERT model. 
Hard voting is used as ensemble method in which the predictions for each BERT experiments are involved in voting to get the final prediction.

Table \ref{tab:bert_ensemble} shows the results of the ensemble models of BERT with different number of experiments each with different random seed.

In the ensemble of four and six seeds when the number of votes is equal, high priority was given to the experiments with the best public score.

The result of the ensemble of four seeds has the best public score, so it was chosen for the final evaluation and got the first place. Although other seeds results had lower public scores, they had higher private scores than the official private score. So actually, our best result is 96.232\% while the official best result is 95.924\%.

\vspace{0.25in}
\section{Conclusion}
This paper describes our participation in NSURL Task 8; Semantic Question Similarity in Arabic.
Different models were proposed for the task; RNN model, CNN model, Multi-head model, BERT model, and ensemble model of BERT. The ensemble model clearly outperforms all other models in this task by achieving 95.924\% F1-Score. This performance ranks first place among nine participating teams.

\bibliography{references}
\bibliographystyle{acl_natbib}

\end{document}